\documentclass{article}
\usepackage[preprint]{corl_2024}

\usepackage{epsfig} 
\usepackage{mathptmx} 
\usepackage{times} 
\usepackage{amssymb}  
\usepackage{caption}
\usepackage{graphicx}
\usepackage{algorithm}
\usepackage{tcolorbox}
\usepackage{wrapfig}
\usepackage{subcaption}
\usepackage{amsmath}
\usepackage{booktabs}
\usepackage{multirow}
\usepackage{hyperref}

\newcommand{\llm}{VLM-planner}
\newcommand{\tpvqa}{\textsc{DKPrompt}}
\newcommand{\openloop}{Classical-planner}
\newcommand{\success}{Suc.-QA}
\newcommand{\affordance}{Aff.-QA}
\newcommand{\successaffordance}{Suc.Aff.-QA}
\newcommand{\effect}{Eff.-only}
\newcommand{\precondition}{Pre.-only}

\usepackage{listings}

\lstdefinelanguage{PDDL}{
    morekeywords={define,domain,requirements,strips,typing,types,predicates,action,parameters,precondition,effect,and,not,or,forall,when,:either,:domain,:requirements,:types,:predicates,:action,:parameters,:precondition,:effect},
    morecomment=[l]{;},
    morecomment=[s]{/*}{*/},
    morestring=[b]",
}

\title{\LARGE \bf
\tpvqa{}: Domain Knowledge Prompting Vision-Language Models for Open-World Planning
}

\author{
  \begin{tabular}{c@{\hspace{1em}}c@{\hspace{1em}}c@{\hspace{1em}}c@{\hspace{1em}}c}
   Xiaohan Zhang$^{1}$ & Zainab Altaweel$^{1}$\thanks{~indicates equal contribution.}  & Yohei Hayamizu$^{1*}$ & Yan Ding$^{1}$ & Saeid Amiri$^{1}$\\
   \end{tabular}\\
  \begin{tabular}{c@{\hspace{1em}}c@{\hspace{1em}}c@{\hspace{1em}}c}
   \textbf{Hao Yang}$^{2}$    & 
   \textbf{Andy Kaminski}$^{2}$  & 
   \textbf{Chad Esselink}$^{2}$ &
   \textbf{Shiqi Zhang}$^{1}$\\[5pt]
  \end{tabular}\\
  \begin{tabular}{c@{\hspace{1em}}c}
  \textnormal{$^{1}$State Univeristy of New York at Binghamton}            & 
  \textnormal{$^{2}$Ford Research}
\end{tabular} \\[5pt]
}

\begin{document}
\maketitle

\thispagestyle{empty}
\pagestyle{empty}

\begin{abstract}
Vision-language models (VLMs) have been applied to robot task planning problems, where the robot receives a task in natural language and generates plans based on visual inputs. 
While current VLMs have demonstrated strong vision-language understanding capabilities, their performance is still far from being satisfactory in planning tasks. 
At the same time, although classical task planners, such as PDDL-based, are strong in planning for long-horizon tasks, they do not work well in open worlds where unforeseen situations are common. 
In this paper, we propose a novel task planning and execution framework, called \tpvqa, which automates VLM prompting using domain knowledge in PDDL for classical planning in open worlds. 
Results from quantitative experiments show that \tpvqa ~outperforms classical planning, pure VLM-based and a few other competitive baselines in task completion rate.~\footnote{\url{https://dkprompt.github.io/}}
\end{abstract}


\section{Introduction}
Prompting foundation models such as large language models (LLMs) and vision-language models~(VLMs) requires extensive domain knowledge and manual efforts, resulting in the so-called ``prompt engineering'' problem. 
To improve the performance of foundation models, one can provide examples explicitly~\cite{brown2020language} or implicitly~\cite{lester2021power}, or encourage intermediate reasoning steps~\cite{wei2022chain, yao2024tree}. 
Despite all the efforts, their performance in long-horizon reasoning tasks is still limited. 
Classical planning methods, including those defined by Planning Domain Definition Language~(PDDL), are strong in ensuring the soundness, completeness and efficiency in planning tasks~\cite{ghallab2016automated}.
However, those classical planners rely on predefined states and actions, and do not perform well in open-world scenarios. 
We aim to enjoy the openness of VLMs in scene understanding while retaining the strong long-horizon reasoning capabilities of classical planners. 
Our key idea is to extract domain knowledge from classical planners for prompting VLMs towards enabling classical planners that are visually grounded and responsive to open-world situations. 

Given the natural connection between planning symbols and human language, this paper investigates how pre-trained VLMs can assist the robot in realizing symbolic plans generated by classical planners, while avoiding the engineering efforts of checking the outcomes of each action. 
Specifically, we propose a novel task planning and execution framework called \tpvqa{}, which prompts VLMs using domain knowledge in PDDL, generating visually grounded task planning and situation handling.
\tpvqa{} leverages VLMs to detect action failures and verify action affordances towards successful plan execution. 
We take the advantage of the domain knowledge encoded in classical planners, including the actions defined by their effects and preconditions.
By simply querying current observations against the action knowledge, similar to applying VLMs to Visual Question Answering~(VQA) tasks, \tpvqa{} can trigger the robot to address novel situations and recover from action failures.

\begin{figure}[t]
\begin{center}
    \includegraphics[width=0.72\textwidth]{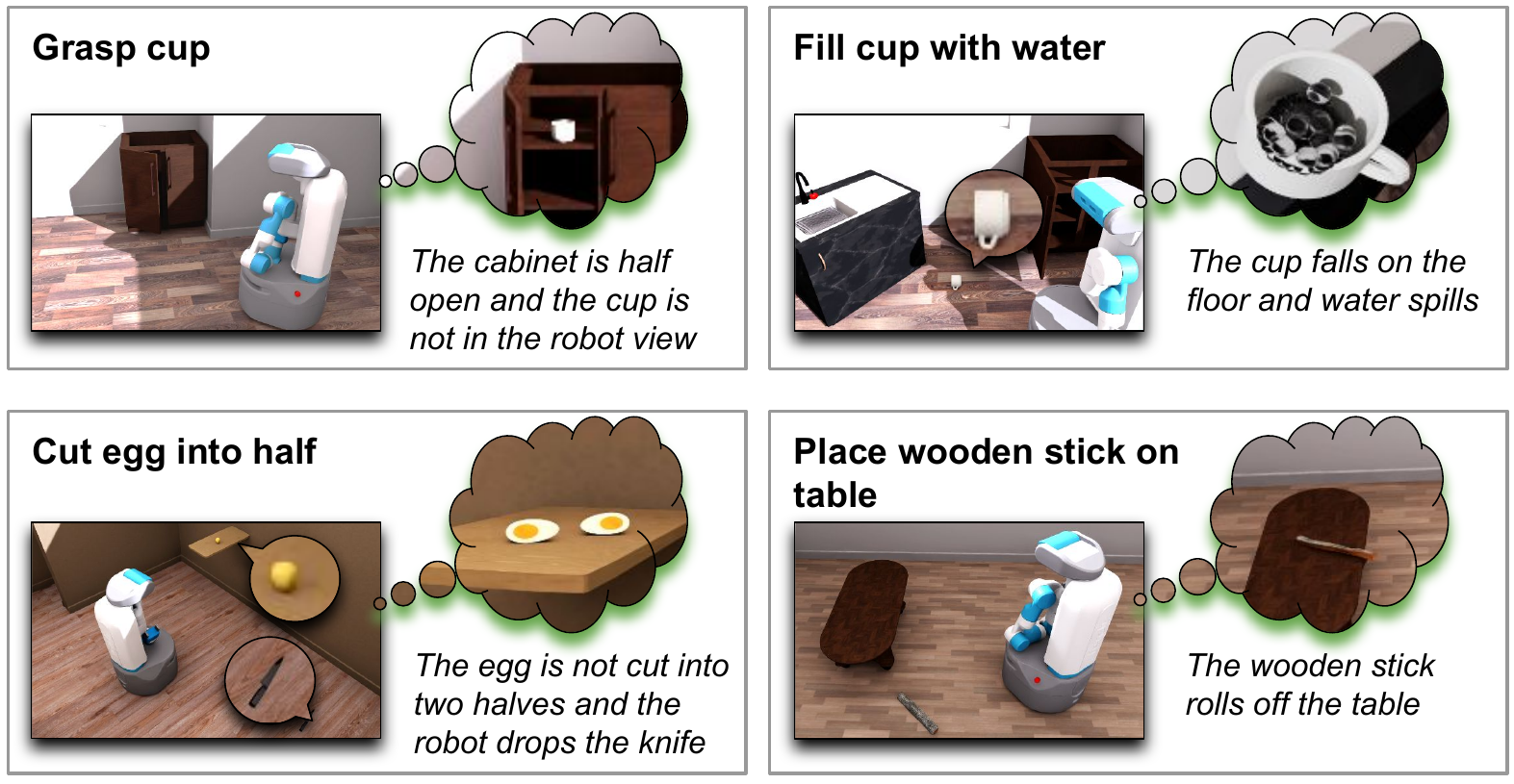}
    \caption{A few unforeseen situations during action execution. In the top-left example, the robot ``opened'' the cabinet door to get prepared for grasping the cup. It was expected that the cup in white would have been in the robot's view after the ``opening'' action, while a situation occurred, i.e., the cabinet was only half-open. \tpvqa{} prompts vision-language models (VLMs) using domain knowledge to detect and address such situations. While one can develop a safeguard to detect cabinet opening being successful, our goal is to automate this process, avoiding such manual efforts and handling unforeseen situations.  
    }
    \vspace{-2em}
\label{fig:teaser}
\end{center}
\end{figure}

We conducted quantitative evaluations using the OmniGibson simulator~\cite{li2023behavior}. 
We assume that robot actions are \textit{imperfect} by nature, frequently causing \textit{situations}\footnote{Situation is an unforeseen world state that potentially prevents an agent from completing a task using a solution that
normally works~\cite{ding2023integrating}.} during execution (Figure~\ref{fig:teaser}).
Results demonstrate that \tpvqa{} utilizes domain knowledge to adaptively generate task plans, recovers from action failures and re-plans when situations occur.
In addition, we hope that researchers working on VLMs, robot planning or both find our evaluation platform useful for their research. 
In particular, the open-world situations and structured world knowledge presents a new playground for comparing robot planning and vision-language understanding using large-scale models.

\section{Related Work}
This section starts with covering a wide range of downstream applications of classical AI planners in symbolic task planning.
It then explores the role of Large Language Models (LLMs) in robot planning, discussing their strengths (e.g., rich in commonsense) and limitations (e.g., lack of correctness guarantee). 
Finally, it examines the recent advancements in Vision-language Models (VLMs) and their impact on the robotics community.

\subsection{Classical AI Planning for Robots}
Automated planning algorithms have a long-standing history in the literature of symbolic AI and have been widely used in robot systems. 
Shakey is the first robot that was equipped with a planning component, which was constructed using STRIPS~\cite{nilsson1984shakey}. 
Recent classical planning systems designed for robotics commonly employ Planning Domain Description Language (PDDL) or Answer Set Programming (ASP) as the underlying action language for planners~\cite{jiang2019task, brewka2011answer, lifschitz2002answer, fox2003pddl2, lagriffoul2018platform, kaelbling2013integrated, zhang2015mobile, ding2020task, jiang2019multi, ding2022learning}.
Most classical planning algorithms that are designed for robot planning do not consider perception.
Though some recent works have already shown that training vision-based models from robot sensory data can be effective in plan feasibility evaluation~\cite{zhu2021hierarchical, zhang2022visually, driess2020deep, driess2020deeph, wells2019learning}, their methods did not tightly bond with language symbols which are the state representations for classical planning systems.  
The most relevant work to our study is probably the research by~\citeauthor{migimatsu2022grounding}, which trained domain-specific predicate classifiers from webscale data and deployed on a robot planning system~\cite{migimatsu2022grounding}.
We propose \tpvqa{} that investigates how off-the-shelf Vision-language Models connect perception with symbolic language which is used to represent robot knowledge.

\begin{figure*}
\begin{center}
    \includegraphics[width=\textwidth]{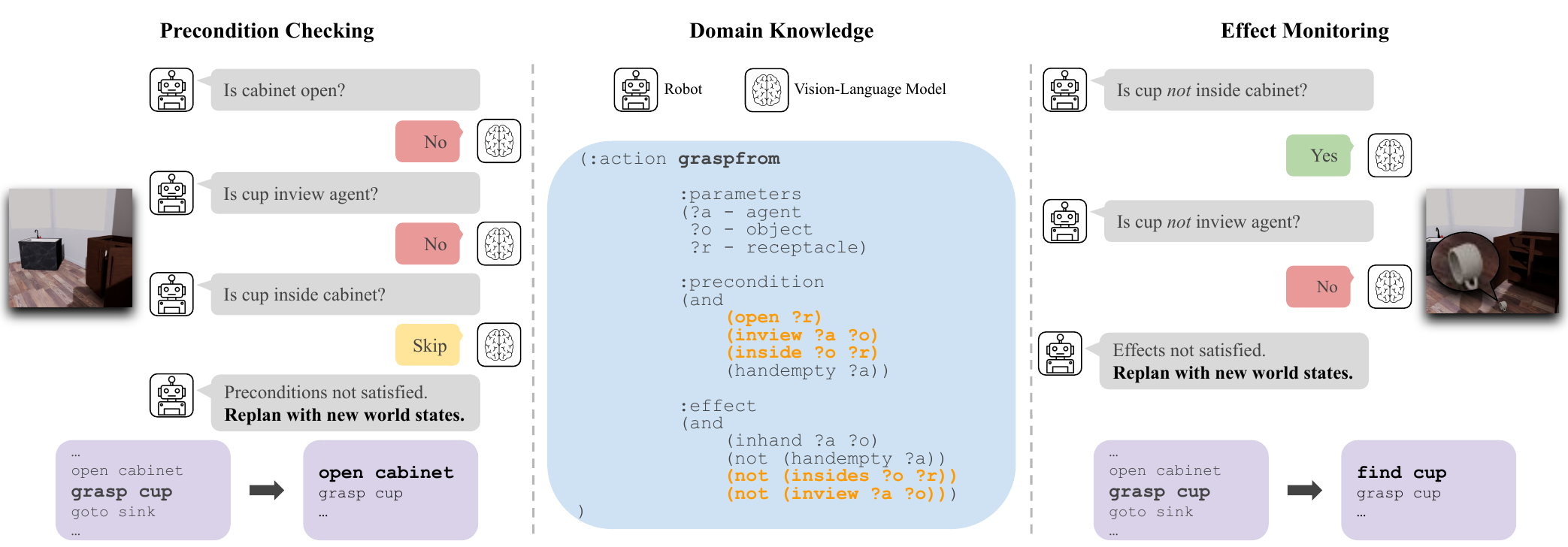}
    \caption{An overview of \tpvqa{}. By simply querying the robot's current observation against the domain knowledge~(i.e., action preconditions and effects) as VQA tasks, \tpvqa{} can call the classical planner to generate a new valid plan using updated world states. Note that \tpvqa{} only queries about predicates. The left shows how \tpvqa{} checks every precondition of the action to be executed next, and the right shows how it verifies the expected action effects are all in place after action execution. Replanning is triggered when preconditions or effects are unsatisfied after updating the planner's action knowledge. 
    }
    \vspace{-2em}
    \label{fig:overview}
\end{center}
\end{figure*}
\subsection{Classical Planning with Large Language Models}
In the light of the recent advancement in artificial intelligence, many LLMs have been developed in recent years~\cite{devlin2018bert, openai, chen2021evaluating, zhang2022opt}.
These LLMs can encode a large amount of common sense~\cite{liu2023pre} and have been widely applied to robot task planning~\cite{kant2022housekeep,huang2022language,ahn2022can,huang2022inner,singh2022progprompt,zhao2023large,liu2022structdiffusion,wu2023tidybot,rana2023sayplan}. 
However, a major drawback of existing LLMs is their lack of long-horizon reasoning/planning abilities for complex tasks~\cite{valmeekam2022large, valmeekam2023planning, openai2023gpt4}. 
Specifically, the output they produce when presented with such a task is often incorrect in the sense that following the output plan will not actually solve the task.  
As a result, a wide range of studies have investigated approaches that combine the classical planning methodology with LLMs in robotic domains~\cite{silver2022pddl, pallagani2022plansformer, arora2023learning, silver2024generalized, chen2023autotamp, wang2024llm, liu2023llm+, stein2023autoplanbench, guan2023leveraging, ding2023task}.
However, neither LLMs nor classical planners are inherently \textit{grounded}, often necessitating complex interfaces to bridge the symbolic-continuous gap between language and robot perception. 
Our approach seeks to ground classical planners by utilizing pre-trained VLMs through a novel but straightforward domain knowledge prompting strategy.

\subsection{Vision-language Models in Robotics}
VLMs have emerged as powerful methods that integrate visual and linguistic information for complex AI tasks~\cite{zhang2024mm, radford2021learning, achiam2023gpt, team2023gemini, claude3}. 
Researchers have started to employ such models in robot systems~\cite{wake2023gpt, lykov2024cognitivedog, guan2024task, majumdar2024openeqa, sermanet2023robovqa}, where these models have shown effectiveness in, for example, semantic scene understanding~\cite{ha2022semantic}, open-ended agent learning~\cite{fan2022minedojo}, guiding robot navigation~\cite{shafiullah2022clipfields} and manipulation behaviors~\cite{shridhar2021cliport, moo2023arxiv}.
Recent VLMs have also been used for building \textit{planning} frameworks~\cite{lv2024robomp2, zhao2023chat}.
Adaptive planning significantly improve task performance through better environment awareness and fault recovery, and language understanding allows robots to seek human assistance in handling uncertainty~\cite{ren2023robots, zhi2024closedloop}. 
There have been recent methods, similar to us, that query VLMs for action success, failures, and affordances~\cite{du2023vision, driess2023palm, guo2023doremi}. 
Different from their work, \tpvqa{} uses classical planners to generate executable symbolic plans rather than solely relying on pre-trained models.
Additionally, \tpvqa{} integrates domain knowledge into the prompts, enhancing the grounded connection between the VLMs and the symbolic planner.

\section{\tpvqa{} for Planning in Open Worlds}
\label{sec:method}

This section presents the implementation details of \tpvqa{} in a robot planning system, particularly suitable for open worlds.
The system assumes that the agent is equipped with a predefined set of actions which are \textit{imperfect} by nature, frequently causing unforeseen situations (Section~\ref{sec:actions}).
The agent also processes a handful of \textit{action knowledge}, where actions are defined by their preconditions and effects.
These preconditions and effects are further represented as \textit{objects} and \textit{propositions}, i.e., predicates (Section~\ref{sec:preds}).
We then introduce how \tpvqa{} takes advantages of the action knowledge for states update and online re-planning (Section~\ref{sec:plan}).

\subsection{Robot Actions}
\label{sec:actions}
\begin{table*}
\centering
\vspace{-1em}
\caption{Actions, constraints, and their uncertain outcomes (i.e., situations).}
\scriptsize
\renewcommand{\arraystretch}{0.88}
\begin{tabular}{@{}lll@{}}
\toprule
        \cmidrule(rl){2-3}
        Actions & Constraints &
        \def\arraystretch{0.95}Situations
        \def\arraystretch{0.95}\\ \midrule
	find & \begin{tabular}{@{}l@{}}(1) The object and the agent\\are in the same room.\end{tabular} & \begin{tabular}{@{}l@{}}(1) The robot succeeds in navigation but the object is not inview. \\(2) There is no free space near the object so navigation fails. \\(3) The object that the robot is holding drops during navigation. \end{tabular} \\ \midrule
	  grasp & \begin{tabular}{@{}l@{}}(1) The object is inview.\\(2) The agent's hand is empty.\end{tabular} & \begin{tabular}{@{}l@{}}(1) The robot fails to grasp, and the object position remains unchanged.\\(2) The robot fails to grasp, and the object drops nearby. \end{tabular}\\ \midrule
	  placein & \begin{tabular}{@{}l@{}}(1) The object is inhand.\\(2) The receptacle is inview.\\(3) The receptacle is not closed. \end{tabular} & \begin{tabular}{@{}l@{}}(1) The robot fails to place, and the object remains in the robot's hand.\\(2) The robot fails to place, and the object drops nearby. \end{tabular}\\ \midrule
	  placeon & \begin{tabular}{@{}l@{}}(1) The object is inhand.\\(2) The receptacle is inview.\end{tabular} & \begin{tabular}{@{}l@{}}(1) The robot fails to place, and the object remains in the robot's hand.\\(2) The robot fails to place, and the object drops nearby.\end{tabular}\\ \midrule
	  fillsink & \begin{tabular}{@{}l@{}}(1) The sink is inview.\end{tabular} & \begin{tabular}{@{}l@{}}(1) The robot fails to open the faucet. \end{tabular}\\\midrule
        fill & \begin{tabular}{@{}l@{}}(1) The container is inhand.\\(2) The agent is near sink. \\(3) The container is empty. \end{tabular} & \begin{tabular}{@{}l@{}}(1) The container is not fully filled.\\(2) The container drops nearby. \end{tabular}\\ \midrule
        open & \begin{tabular}{@{}l@{}}(1) The object is inview.\end{tabular} & \begin{tabular}{@{}l@{}}(1) The robot fails to open and the object remains closed. \end{tabular}\\ \midrule
        close & \begin{tabular}{@{}l@{}}(1) The object is inview.\end{tabular} & \begin{tabular}{@{}l@{}}(1) The robot fails to close and the object remains open.\end{tabular} \\ \midrule
        turnon & \begin{tabular}{@{}l@{}}(1) The object is inview.\end{tabular} & \begin{tabular}{@{}l@{}}(1) The robot fails to turn on the switch and the object remains off.\end{tabular}\\ \midrule
        cut & \begin{tabular}{@{}l@{}}(1) The object is inview.\\(2) A knife is inhand. \end{tabular} & \begin{tabular}{@{}l@{}}(1) The object is not cut into half, and the knife is still in the robot's hand.\\(2) The object is not cut into half, and the knife drops nearby.\end{tabular}\\
\bottomrule
\label{tab:situations}
\vspace{-2em}
\end{tabular}
\end{table*}
Our system considers ten actions (as listed in Table~\ref{tab:situations}), including basic navigation and manipulation.
Situations occur after actions are successfully triggered by the agent.
Table~\ref{tab:situations} also provides examples of situations that happen following specific actions.
Some of these situations impact the world states, while others do not.
For example, the robot may fail on a ``grasp'' action, resulting in the target object, originally on the table, to fall on the floor nearby (changing the state from \texttt{on(obj, table)} to \texttt{on(obj, floor)}).
On the other hand, the object might also remain on the table with the world states being unchanged.
To quantify the openness of different environments, we created the simulation platform in such a way that one can easily adjust the probability of a situation's occurrence. 
The source code of our benchmark system will be made available in our project website. 

Actions are implemented in a discrete manner for simplification purposes, since continuous action execution is not this paper's focus.
For instance, ``find'' action is implemented by teleporting the agent from its initial position to a randomly-sampled obstacle-free goal position near the target, and ``fill'' action is by adding fluid particles directly into the container that the robot is holding.

Actions are subject to several constraints.
For example, ``grasp'' action is deemed executable only if the target object is in the agent's view (assuming vision-based manipulation) and the agent's hand is empty.
Similarly, ``cut'' action is considered executable only if the object to be cut is in the agent's view and the agent is currently holding a knife.
Calling an action with at least one unsatisfied constraint will result in an action failure, but without any changes to the world states.
Note that such constraints are not made available to agents, instead, they are partially encoded as domain (action) knowledge that the agent possesses.

We assume that situations can only happen during action execution, but are only observable by agents either before or after the action execution phase. 
This assumption indicates that situations are solely caused by actions, and we are aware of a few recent robotic research that have started to consider more generalized situation handling~\cite{ding2023integrating}. 
We leave situations that caused by external environmental factors (human or other embodiments) to future work.

\subsection{Predicates}
\begin{table*}
\centering
\scriptsize
\caption{\tpvqa{} assumptions for predicates.}
\begin{tabular}{@{}lc@{}}
\toprule
	\bf Perceptible in vision & inview, closed, open, inside, halved, onfloor, ontop, cooked\\ \midrule
	  \bf Perceptible in non-vision & handempty, inhand, hot\\ \midrule
	  \bf Imperceptible & turnedon, filled, inroom, insource\\
\bottomrule
\label{tab:predicates}
\vspace{-2em}
\end{tabular}
\end{table*}
\label{sec:preds}
A single action is usually defined by multiple preconditions and effects in the domain knowledge.
VLMs, especially for those that are not trained using domain-specific data, frequently produce inaccurate answers that cause disagreements among the given preconditions~(or effects).
For instance, the VLM might answer ``Yes'' to both \texttt{on(apple, table)} and \texttt{inhand(apple)} after the robot picks up an apple from the table.
In this paper, \tpvqa{} categorizes predicates into three: \textit{perceptible in vision}, \textit{perceptible in non-vision}, and \textit{imperceptible}.
\tpvqa{} will only ask about \textit{``perceptible in vision''} predicates.
Intuitively, we believe VLMs should be and will be only good at visually-perceptible predicates. 
The robot will then have ground truth access to \textit{perceptible in non-vision} predicates (this assumption also applies to all other baselines).
We leave identifying these predicates using more advanced Multimodal Language Models to future work.
As for the remaining \textit{imperceptible} predicates, the \tpvqa{} agent maintains a positive attitude and assumes they are always True. 
This suggests that \tpvqa{} believes these predicates will never be affected by any situation.

\subsection{\tpvqa{}}
\label{sec:plan}
Before every action execution, \tpvqa ~extracts knowledge about action preconditions from the planner's domain description.
For instance, as indicated in Figure~\ref{fig:overview}, action \texttt{graspfrom($a$, $o$, $r$)} has preconditions of \texttt{open($r$)}, \texttt{inview($a$, $o$)}, \texttt{inside($o$, $r$)}, and \texttt{handempty($a$)}, meaning that to grasp an object $o$ from a receptacle $r$, $r$ should be open (not closed), $o$ should be in the agent's current first person view, $o$ should be inside $r$, and the agent's hand should be empty.
Then, we simply convert each action precondition into a natural language query by using manually defined templates, though it has been evident that LLMs can be used for the translation between PDDL and natural language~\cite{liu2023llm+}. Examples include \textit{``Is \textless{}$o$\textgreater{} inview agent?''} and \textit{``Is \textless{}$o$\textgreater{} inside \textless{}$r$\textgreater{}?''}
Paring each natural language query with the current observation from the robot's first-person view, we call the VLM to get answers indicating if the precondition is satisfied.

According to the results (i.e., ``yes'', ``no'', or ``skip'' if unsure) from the VLM, \tpvqa{} will update the current state information in the classical planning system.
Figure~\ref{fig:overview} (Left) shows an example where the robot wants to \texttt{graspfrom(cup, cabinet)} but fails to detect ``cabinet is open'', ``cup is inview of agent'', and is suspecious about if ``cup is in the cabinet'' (the VLM answers ``skip'' to this question) given the current observation.
As a result, \tpvqa{} will update the current state by changing \texttt{open(cabinet)} to \texttt{closed(cabinet)}, and removing \texttt{inview(agent, cup)}.
\texttt{inside(cup, cabinet)} will remain the same because we do not update the state if the VLM answers ``skip'', indicating the agent holds a positive attitude that situations will not commonly occur.
We then provide the updated world state to the classical planner as the ``new'' initial state to re-generate a plan. 
In the above example, instead of \texttt{graspfrom(cup, cabinet)}, the robot will now take the action of \texttt{open(cabinet)} again according to the newly-generated action plan.
After every action execution, \tpvqa{} extracts knowledge about action effects from the planner's domain description, illustrated in Figure~\ref{fig:overview} (Right). 
It queries action effects by using the VLM. 
If the effects are not satisfied, the robot will update its belief on the current states and re-plan accordingly.
The knowledge-based automated prompting strategy of VLMs enables our planning system to adaptively capture and handle unforeseen situations at execution time. 

\section{Experiments}

We conducted extensive experiments to evaluate the performance of \tpvqa{} comparing with baselines from the literature.
Our hypothesis is \tpvqa{} produces the highest task completion rate because of its effectiveness in plan monitoring and online re-planning using domain knowledge and perception.

\subsection{Experiment Setup}
\begin{table*}
\renewcommand{\arraystretch}{0.85}
\centering
\vspace{-1em}
\caption{Task descriptions and initial plan length.}
\scriptsize
\resizebox{0.85\textwidth}{!}{\begin{tabular}{@{}lcc@{}}
	\toprule
	 Name & Descriptions & Initial plan length\\ \midrule
	 \begin{tabular}{@{}l@{}}boil water in\\the microwave\end{tabular} & Pick up an empty cup in a closed cabinet, fill it with water using a sink, and boil it in a microwave. & 12\\ \midrule
      \begin{tabular}{@{}l@{}}bring in\\empty bottle\end{tabular} & Find two empty bottles in the garden and bring them inside. & 8\\ \midrule
	 \begin{tabular}{@{}l@{}}cook a\\frozen pie\end{tabular} & Take an apple pie out of the fridge and heat it using an oven. & 8 \\ \midrule
	 \begin{tabular}{@{}l@{}}halve\\an egg\end{tabular} & Find a knife in the kitchen and use it to cut a hard-boiled egg into half. & 4\\ \midrule
	 \begin{tabular}{@{}l@{}}store\\firewood\end{tabular} & Collect two wooden sticks and place them on a table. & 8\\
\bottomrule
\label{tab:tasks}
\vspace{-1.5em}
\end{tabular}}
\end{table*}
Quantitative evaluation results are collected in the OmniGibson simulator~\cite{li2023behavior}. 
The agent is equipped with a set of skills, and aims to use its skills to interact with the environment, completing long-horizon tasks autonomously.
In the experiment, we consider five everyday tasks that are ``boil water in the microwave'', ``bring in empty bottle'',``cook a frozen pie'', ``halve an egg'', and ``store firewood''.
Their detailed descriptions are shown in Table~\ref{tab:tasks}.
These five tasks are originally from the Behavior 1K benchmark~\cite{li2023behavior} that are accompanied with the simulator.
Task descriptions including initial and goal states are written in PDDL and symbolic plans are generated using the fast-downward planner~\cite{helmert2006fast}.

\subsection{Results}

\paragraph{Comparisons with Baselines.}
\begin{figure*}[h!]
\begin{center}
\vspace{-1em}
    \includegraphics[width=0.9\textwidth]{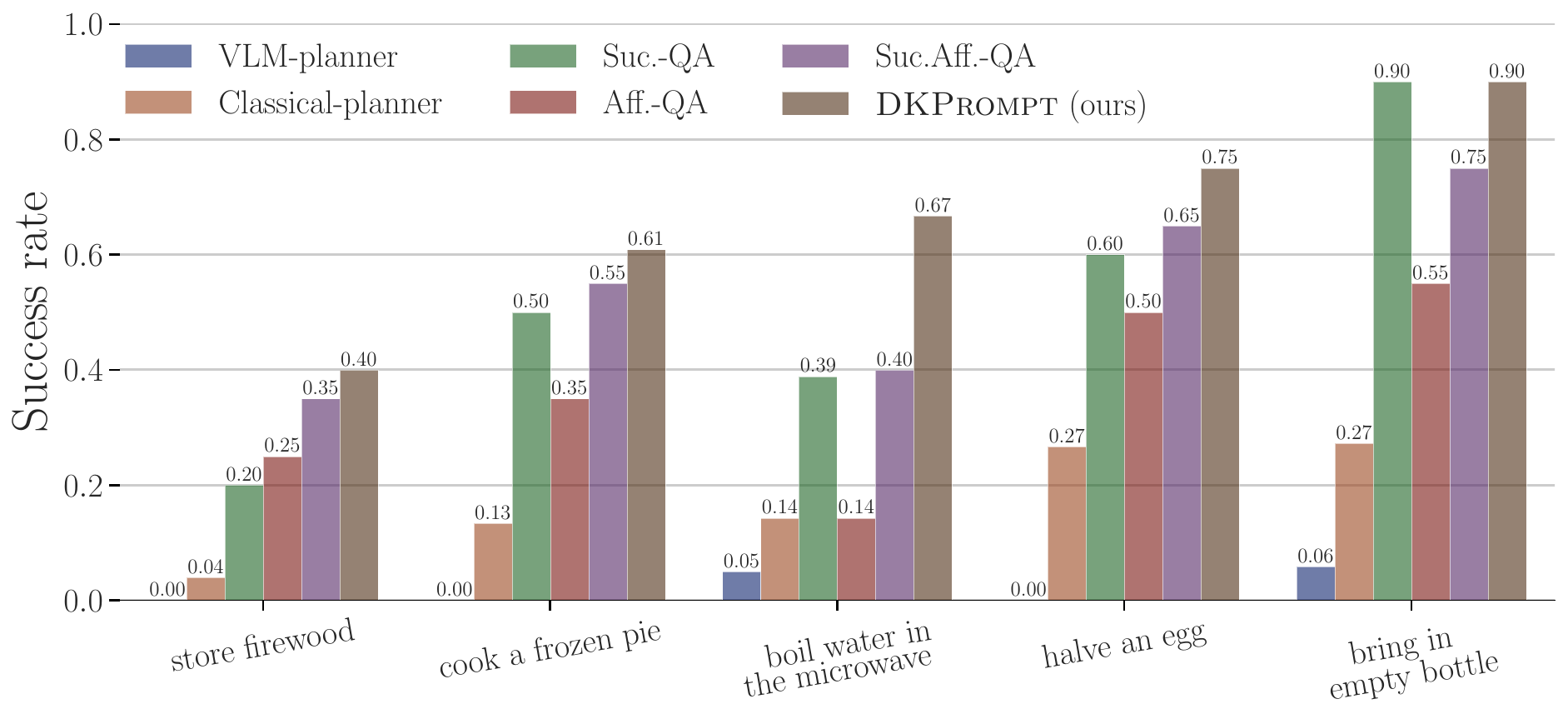}
    \vspace{-.5em}
    \caption{\tpvqa{} v.s. baselines in success rate over five everyday tasks.
    }
\label{fig:baselines}
    \vspace{-1em}
\end{center}
\end{figure*}
Figure~\ref{fig:baselines} presents the main experimental results and details the comparative success rates of various methods from the literature.
The methods include:
\begin{itemize}
    \item \llm{}, which uses the VLM as a planner to generate task plans, similar to~\cite{huang2022language}. For fair comparisons, we also provide domain knowledge (as natural language) in the prompts for the VLM.

    \item \openloop{}, which is a typical classical planning approach without perception, assuming all action executions are successful;
    
    \item  \success{}, which uses a classical planner to generate plans, and asks about action success after each action execution. 
    This baseline is inspired by~\cite{du2023vision}, and we use the same query provided in their paper, which is \textit{``Did the robot successfully $<$action$>$?''} 
    \success{} does not consider if the next action is executable;
    
    \item \affordance{}, which uses a classical planner to generate plans, and asks about action affordance before each action execution. 
    This baseline is designed with prompts provided in the original PaLM-E paper~\cite{driess2023palm}, which are \textit{``Is it possible to $<$action$>$ here?''} and \textit{``Was $<$action$>$ successful?''}
    \affordance{} does not consider whether the previous action is successful;
    
    \item \successaffordance{}, which uses a classical planner, asks about both action affordance (before each action execution) and action success (after each action execution), similar to~\cite{huang2022inner}. 
     \footnote{We use the same VLM as ours (GPT4) for implementing all baselines that require a VLM.}
\end{itemize}
When using VLM itself as the planner, the agent frequently fails in finding an executable plan, resulting in the lowest success rate.
This finding is consistent with recent work~\cite{valmeekam2022large} and motivates the development of other research that combines classical planning with large models~\cite{liu2023llm+}. 
\openloop{}, which operates without visual feedback during task execution, shows the second lowest success rate across five tasks compared to other evaluated methods, highlighting its limited effectiveness in handling situations and recovering from potential action failures. 
In contrast, methods that involve querying for action affordances, success probabilities, or both, acheive much higher success rates as compared to the ``blind'' classical planning approach. 
This improvement demonstrates the general advantage of incorporating visual feedback and high-level reasoning in task planning systems.
While it is always a good practice to verify both before and after an action (like \successaffordance{}), we found that \success{} also surpasses the performance of \affordance{}, indicating that there is a greater positive impact on task completion from action failure recovery, and VLMs have better zero-shot reasoning capabilities on the direct effects caused by actions. 

\begin{wrapfigure}{r}{0.42\textwidth}
  \centering
  \vspace{-1.8em}
  \includegraphics[width=0.42\textwidth]{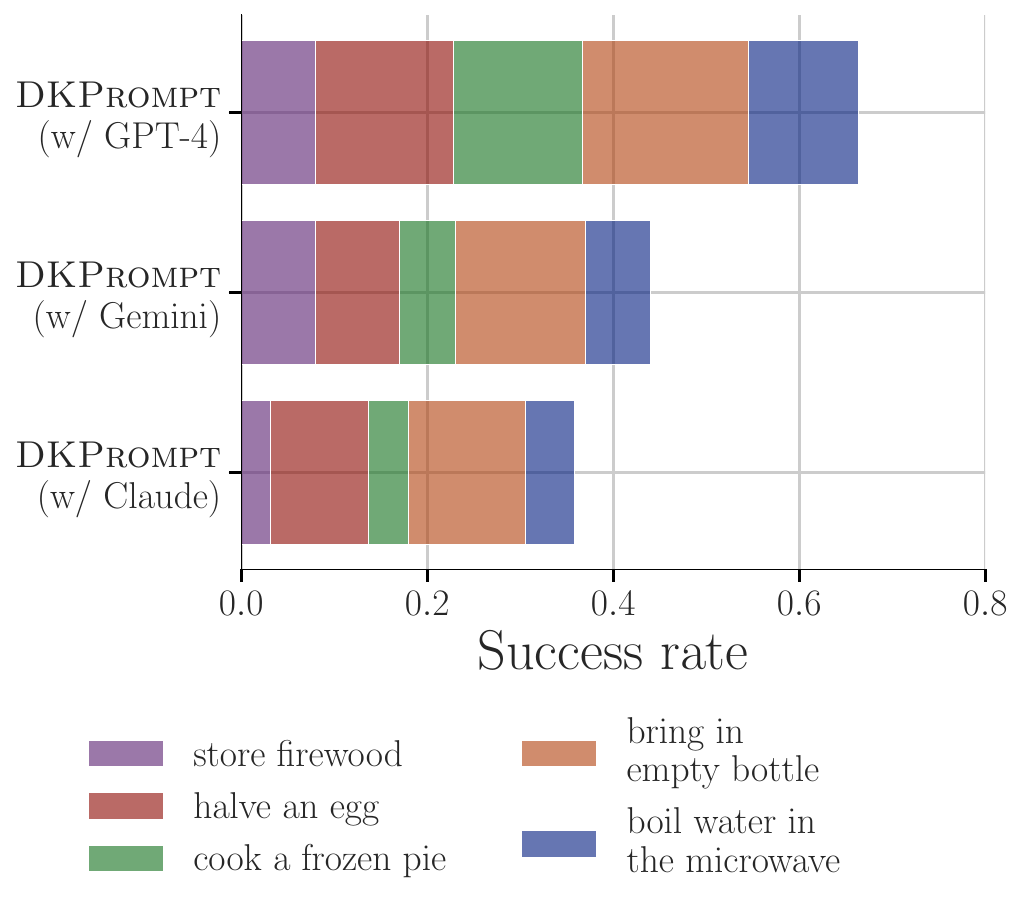}
  \vspace{-1.7em}
  \caption{Performance of off-the-shelf VLMs.}
    \label{fig:vlms}
    \vspace{-3em}
\end{wrapfigure}
We observed that \tpvqa{} consistently outperforms baselines in task completion rates, which supports our hypothesis.
By incorporating domain knowledge~(i.e., action preconditions and effects) for prompting, \tpvqa{} is significantly better than other methods, including \successaffordance{} that also cares about affordance prediction and failure detection.
However, \successaffordance{} queries about actions solely by their names, which provides less information than the detailed domain knowledge used by \tpvqa{}, indicating that action knowledge is more informative for pretrained VLMs to reason over.

\paragraph{Ablation Study on Preconditions and Effects.}
\begin{table*}[t]
    \centering
    \scriptsize
    \renewcommand{\arraystretch}{0.9}
    \vspace{-.5em}
    \caption{Ablation study on preconditions and effects.
    }
    \resizebox{0.8\textwidth}{!}{%
    \begin{tabular}{c@{\hspace{1ex}} l ccccc r}
        \toprule
        & & \multicolumn{5}{c}{Tasks} \\
        \cmidrule(rl){3-7}
        \textsc{\#} & Methods &
        \def\arraystretch{0.95}\begin{tabular}{@{}c@{}}boil water in\\the microwave\end{tabular} &
        \def\arraystretch{0.95}\begin{tabular}{@{}c@{}}bring in\\empty bottle\end{tabular} &
        \def\arraystretch{0.95}\begin{tabular}{@{}c@{}}cook a\\frozen pie\end{tabular} &
        \def\arraystretch{0.95}\begin{tabular}{@{}c@{}}halve\\an egg\end{tabular} &
        \def\arraystretch{0.95}\begin{tabular}{@{}c@{}}store\\firewood\end{tabular} & avg. (\%)\\
        \midrule
        \multicolumn{1}{l}{{\textbf{Ours}}} & \\
        \textsc{1} & \tpvqa{}  & \textbf{66.7} & 90.0 & \textbf{60.9} & \textbf{75.0} & \textbf{40.0} & \bf 66.5\\
        \midrule
        \multicolumn{1}{l}{\textbf{{Ablation}}} & \\
        \textsc{2} & \effect{} & 50.0 & \textbf{93.8} & 26.7 & 66.7 & 28.0 & 53.0\\
        \textsc{3} & \precondition{} & 17.6 & 75.0 & 35.0 & 55.0 & 20.0 & 41.5\\

        \bottomrule
    \end{tabular}%
    }
    \label{tab:ablation}
    \vspace{-2em}
\end{table*}
Table~\ref{tab:ablation} presents an ablation study comparing the performance of different versions of our approach across the same set of tasks.
\tpvqa{} integrates both action effects and action preconditions, while we are also curious to know how they affect the overall task completion independently. 
\tpvqa{} achieves an average success rate of 66.5\%. 
For ablation methods where only action effects are considered (\effect{}), the average success rate drops to 53.0\%, and for methods considering only preconditions (\precondition{}), it further decreases to 41.5\%. 
This suggests that the integration of both effects and preconditions in \tpvqa{} significantly enhances task performance compared to considering these components separately.

\paragraph{Performance of Other VLMs.}
We also run experiments on various VLMs, including GPT-4 (as being used in the original implementation of \tpvqa{}) from OpenAI~\cite{openai2023gpt4}, Gemini 1.5 from Google~\cite{reid2024gemini}, and Claude 3 from Anthropic.
According to Figure~\ref{fig:vlms}, GPT-4 consistently performs better than Gemini and Claude.
By looking at the highest accuracy among all the VLMs (i.e., less than 65\%), our evaluation benchmark (designed with challenging open-world situations and rich domain knowledge) presents a simulation platform, dataset and success criteria that other researchers working on AI planning, VLMs or both might find useful. 
We will open source the benchmark including software and data to the public after the anonymous review phase. 

\subsection{Real-Robot Deployment}

\begin{figure*}[h]
\begin{center}
    \vspace{-.5em}
    \includegraphics[width=0.8\textwidth]{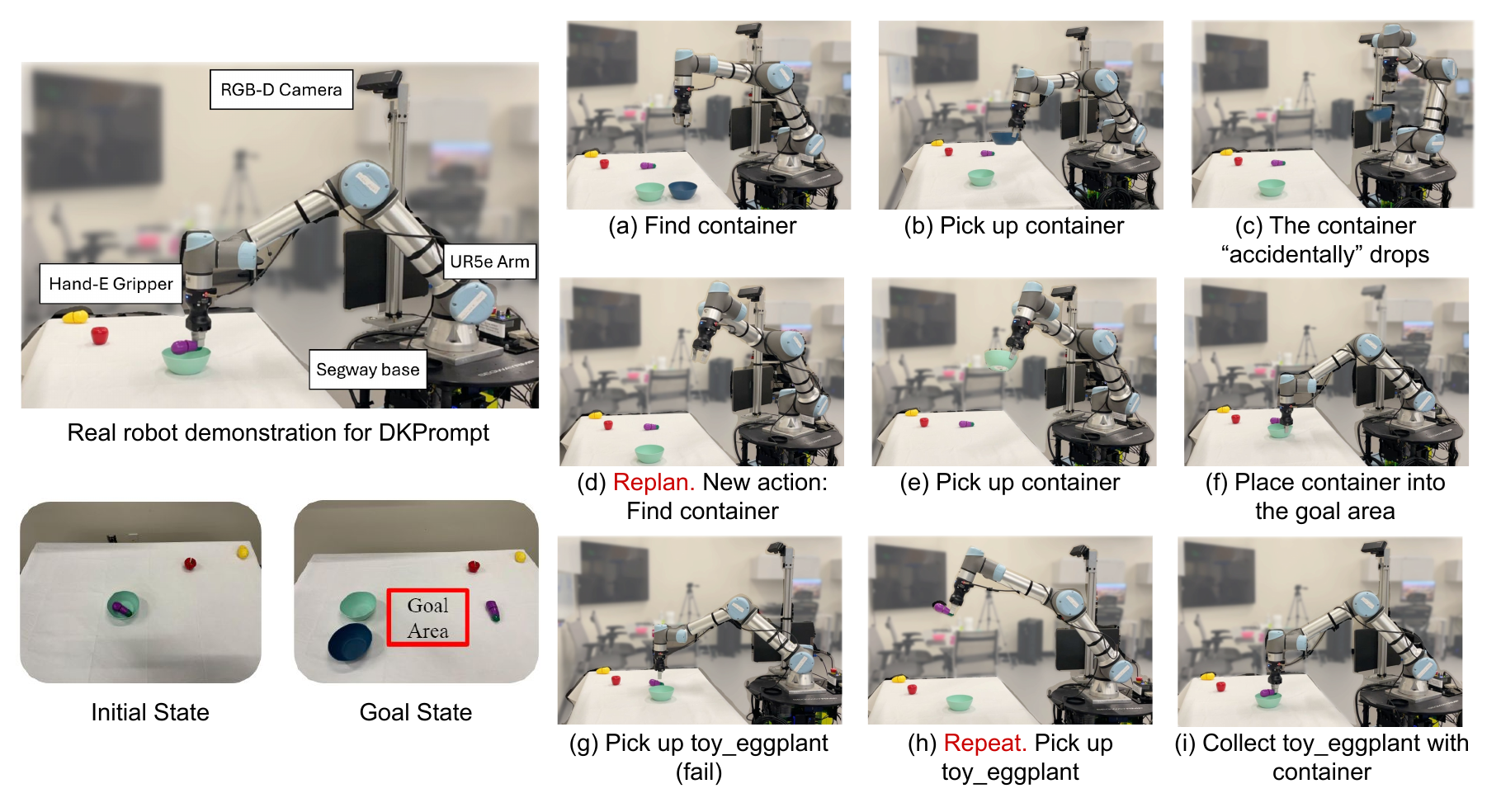}
    \caption{Screenshots showing the full demonstration trial of \tpvqa ~as applied to a real robot.
    }
    \vspace{-1em}
\label{fig:real_long}
\end{center}
\end{figure*}

We also deployed \tpvqa ~on real robot hardware to perform object rearrangement tasks~(Figure~\ref{fig:real_long}), where the goal is to ``collect'' toys using a container and place them in the middle of the table~(i.e., goal area).
Our real-robot setup includes a UR5e Arm with a Hand-E gripper mounted on a Segway base, and an overhead RGB-D camera (relatively fixed to the robot) for perception.
We assume that the robot has a predefined set of skills, including \texttt{pick}, \texttt{place}, and \texttt{find}.
\texttt{Pick} and \texttt{place} actions are implemented using GG-CNN~\cite{morrison2018closing}, and \texttt{find} action simply uses base rotation for capturing tabletop images from different angles.

Given the task description, the robot first decided to execute ``find container'' and ``pick up container''.
These two actions were successfully executed as shown in Figure~\ref{fig:real_long}(a),~\ref{fig:real_long}(b).
When the robot was preparing for the next action~(i.e., ``Place container into the goal area''), the blue container accidentally dropped from the robot's gripper to the ground~(Figure~\ref{fig:real_long}(c)).
Instead of directly executing the next action, \tpvqa ~enabled the robot to check preconditions by querying the VLM \textit{``Is the container in a robot's hand?''}
After receiving negative feedback from the VLM, \tpvqa ~updated the world state by removing \texttt{in\_hand(container)} and called the planner to generate a new plan that started the task again by finding another container~(Figure~\ref{fig:real_long}(d)).
Then the robot picked up the cyan container and placed it in the middle of the table as shown in Figure~\ref{fig:real_long}(e),~\ref{fig:real_long}(f).
The subsequent actions in the plan were to find and pick up a toy, but the \texttt{pick} action failed~(Figure~\ref{fig:real_long}(g)).
\tpvqa ~managed to detect the failure by querying 1) \textit{``Is there a toy\_eggplant on the table?''}, and 2) \textit{``Is the toy\_eggplant in a robot's hand?''}, and receiving Yes and No answers respectively.
As a result, our system suggested the robot repeat the \texttt{pick} action again~(Figure~\ref{fig:real_long}(h)).
Finally, the robot successfully collected the toy by putting it into the cyan container that was previously placed in the goal area~(Figure~\ref{fig:real_long}(i)).

\section{Conclusion}
In this paper, we built the synergy between classical task planning and large Vision-Language Models (VLMs), focusing on how VLMs facilitate robot planning in open-world scenarios.
We propose \tpvqa{} which automates VLM prompting using domain knowledge in PDDL for classical planning and task execution.
Experimental results demonstrate that \tpvqa{} adaptively generate visually-grounded task plans, recovers from action failures and re-plans when situations occur, outperforming classical planning, pure VLM-based and a few other competitive baselines.

\bibliography{ref}
\newpage
\section*{Appendix}


This appendix document presents additional information about our \tpvqa{}~work. 
\tpvqa{}~assists robots in open-world planning tasks by leveraging domain knowledge to automate vision-language model (VLM) prompting. 
In this appendix document, we present our prompt template, domain knowledge in PDDL format, environment settings of open-world planning, and additional experiment results. 
The main goal of this appendix is to improve the reproducibility of this research and we hope robot learning practitioners finds it useful. 

Note that other than this appendix, we have a webpage (\url{https://dkprompt.github.io/}) that serves as a central place where people can find relevant documents about \tpvqa{}.

The following shows the prompt template that was used for querying the VLMs in this research. 
The prompt includes three components. 
The ``System'' part is shared by all prompts for contextualizing the interaction with the VLMs and specifying the format of the output. 
The ``\tpvqa{}'' part lists the questions that are automatically extracted from the domain knowledge in PDDL format.
Finally, each VLM prompt includes an 256x256 image taken from the current robot's observation (``a cup in an open cabinet'' in the following example).

\begin{tcolorbox}[title=\textbf{\tpvqa{}}, colframe=gray, center title]
\begin{wrapfigure}{r}{0.3\textwidth}
  \centering
  \includegraphics[width=0.3\textwidth]{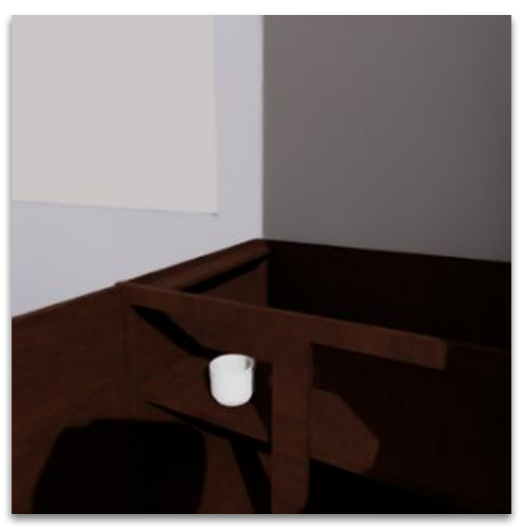}
\end{wrapfigure}
\textbf{System:} Imagine you are an intelligent agent that can answer questions based on what you see.
You will be given a single image as the agent's current view, and one or more yes/no question(s) asking about the image.
Questions will be separated by semicolon.
For each question, you should answer "yes", "no", or "skip" without any explanation.
Answer "yes" or "no" only if you are pretty sure about what you see in the image. 
It's fine to answer "skip" to skip the question if you are not confident about your answer.
Answers should be separated by semicolon (e.g., "yes;no;skip" for three questions).
\newline
\newline
\textbf{\tpvqa{}:} Is cup inview agent?;Is cup inside cabinet?;Is cabinet open?
\end{tcolorbox}

\vspace{1em}
Next we show a series of figures that together form the domain knowledge in PDDL form we provided for the robot to perform task planning and VLM prompting (Figures~\ref{fig:pddl1}-\ref{fig:pddl3}). 
In line with all PDDL-based planning systems, the domain knowledge includes a complete description of the robot's actions (e.g., ``find'' and ``graspon''), where each action is specified by its preconditions and effects.  
Note that we only present the domain description file here. 
A complete planning problem would further require a problem description that includes a description of the current and goal states. 
Since the problem description file changes in each trial, please refer to our github page (link available on the project page shared at the beginning of this appendix) on the instructions of extracting it from the simulator. 

\pagebreak

\begin{figure*}[h!]
\begin{lstlisting}
(define (domain omnigibson)

    (:requirements :strips :typing :negative-preconditions :conditional-effects)

    (:types
        movable liquid furniture room agent - object

        wooden_stick tupperware brownie beer_bottle water_bottle mug pie carving_knife hard__boiled_egg - movable
        water - liquid
        countertop electric_refrigerator oven cabinet sink floor microwave table - furniture

        kitchen living_room - room

        water-n-06 - water
        mug-n-04 - mug
        cabinet-n-01 - cabinet
        sink-n-01 - sink
        floor-n-01 - floor
        microwave-n-02 - microwave
        pie-n-01 - pie
        oven-n-01 - oven
        electric_refrigerator-n-01 - electric_refrigerator
        carving_knife-n-01 - carving_knife
        countertop-n-01 - countertop
        hard__boiled_egg-n-01 - hard__boiled_egg
        water_bottle-n-01 - water_bottle
        beer_bottle-n-01 - beer_bottle
        brownie-n-03 - brownie
        tupperware-n-01 - tupperware
        wooden_stick-n-01 - wooden_stick
        table-n-02 - table

        agent-n-01 - agent
    )

    (:predicates
        (inside ?o1 - object ?o2 - object)
        (insource ?s - sink ?w - liquid)
        (inroom ?o - object ?r - room)
        (inhand ?a - agent ?o - object)
        (inview ?a - agent ?o - object)
        (handempty ?a - agent)
        (closed ?o - object)
        (filled ?o - movable ?w - liquid)
        (filledsink ?s - sink ?w - liquid)
        (turnedon ?o - object)
        (cooked ?o - object)
        (found ?a - agent ?o - object)
        (frozen ?o - object)
        (hot ?o - object)
        (halved ?o - object)
        (onfloor ?o - object ?f - floor)
        (ontop ?o1 - object ?o2 - object)
    )

\end{lstlisting}
\caption{Domain knowledge in PDDL format (Part 1/3). }
\label{fig:pddl1}
\end{figure*}

\begin{figure*}[h!]
\vspace{8em}
\begin{lstlisting}
    (:action find
        :parameters (?a - agent ?o - object ?r - room)
        :precondition (and (inroom ?a ?r) (inroom ?o ?r))
        :effect (and (inview ?a ?o) (found ?a ?o) (forall
                (?oo - object)
                (when
                    (found ?a ?oo)
                    (not (found ?a ?oo)))))
    )

    (:action graspon
        :parameters (?a - agent ?o1 - movable ?o2 - object)
        :precondition (and (inview ?a ?o1) (found ?a ?o1) (handempty ?a) (ontop ?o1 ?o2))
        :effect (and (not (inview ?a ?o1)) (not (handempty ?a)) (inhand ?a ?o1) (not (ontop ?o1 ?o2)))
    )

    (:action graspin
        :parameters (?a - agent ?o1 - movable ?o2 - object)
        :precondition (and (inview ?a ?o1) (found ?a ?o1) (handempty ?a) (inside ?o1 ?o2))
        :effect (and (not (inview ?a ?o1)) (not (handempty ?a)) (inhand ?a ?o1) (not (inside ?o1 ?o2)))
    )

    (:action placein
        :parameters (?a - agent ?o1 - movable ?o2 - object)
        :precondition (and (not (handempty ?a)) (inhand ?a ?o1) (inview ?a ?o2) (found ?a ?o2) (not (closed ?o2)))
        :effect (and (handempty ?a) (not (inhand ?a ?o1)) (inside ?o1 ?o2) (forall
                (?oo - object)
                (when
                    (inside ?oo ?o1)
                    (inside ?oo ?o2))
            ))
    )

    (:action placeon
        :parameters (?a - agent ?o1 - movable ?o2 - object)
        :precondition (and (not (handempty ?a)) (inhand ?a ?o1) (inview ?a ?o2) (found ?a ?o2))
        :effect (and (handempty ?a) (not (inhand ?a ?o1)) (ontop ?o1 ?o2))
    )

    (:action fillsink
        :parameters (?a - agent ?s - sink ?w - liquid)
        :precondition (and (inview ?a ?s) (found ?a ?s) (insource ?s ?w))
        :effect (filledsink ?s ?w)
    )


\end{lstlisting}
\caption{Domain knowledge in PDDL format (Part 2/3).}
\label{fig:pddl2}
\end{figure*}

\pagebreak

\begin{figure*}[h!]
\begin{lstlisting}

    (:action fill
        :parameters (?a - agent ?o - movable ?s - sink ?w - liquid)
        :precondition (and (inhand ?a ?o) (not (handempty ?a)) (filledsink ?s ?w) (inview ?a ?s) (found ?a ?s))
        :effect (and (filled ?o ?w) (not (filledsink ?s ?w)))
    )
    
    (:action openit
        :parameters (?a - agent ?o - object ?r - room)
        :precondition (and (inview ?a ?o) (found ?a ?o) (inroom ?o ?r))
        :effect (and (not (closed ?o)) (forall
                (?oo - object)
                (when
                    (inside ?oo ?o)
                    (inroom ?oo ?r))
            ))
    )
    
    (:action closeit
        :parameters (?a - agent ?o - object ?r - room)
        :precondition (and (inview ?a ?o) (found ?a ?o) (inroom ?o ?r))
        :effect (and (closed ?o) (forall
                (?oo - object)
                (when
                    (inside ?oo ?o)
                    (not (inroom ?oo ?r)))
            ))
    )

    (:action microwave_water
        :parameters (?a - agent ?m - microwave ?o - movable ?w - water)
        :precondition (and (inview ?a ?m) (found ?a ?m) (closed ?m) (inside ?o ?m) (filled ?o ?w))
        :effect (and (turnedon ?m) (cooked ?w))
    )

    (:action heat_food_with_oven
        :parameters (?a - agent ?v - oven ?f - object)
        :precondition (and (inview ?a ?v) (found ?a ?v) (inside ?f ?v))
        :effect (and (hot ?f) (turnedon ?v))
    )

    (:action cut_into_half
        :parameters (?a - agent ?k - carving_knife ?o - object)
        :precondition (and (inview ?a ?o) (found ?a ?o) (not (handempty ?a)) (inhand ?a ?k))
        :effect (halved ?o)
    )

    (:action place_on_floor
        :parameters (?a - agent ?o - object ?f - floor)
        :precondition (and (inview ?a ?f) (found ?a ?f) (not (handempty ?a)) (inhand ?a ?o))
        :effect (and (handempty ?a) (not (inhand ?a ?o)) (onfloor ?o ?f))
    )
)
\end{lstlisting}
\caption{Domain knowledge in PDDL formatf (Part 3/3).}
\label{fig:pddl3}
\end{figure*}

\pagebreak

Next we present Table~\ref{tab:bytask} that includes the complete results of a number of methods, where each column corresponds to a different task. 
The table includes four parts: ours, baselines, ablations and other VLMs. 
The ``Baseline in Literature" part corresponds to the results presented in Figure~\ref{fig:baselines}. 
The ``Ablations'' part corresponds to the results presented in Table~\ref{tab:ablation}. 
The ``Other VLMs" part corresponds to the results presented in Figure~\ref{fig:vlms}. 
Overall, we do not see a huge variance in those methods' performance in different tasks, indicating that our claims about the superiority of \tpvqa{} are valid in different tasks.

\begin{table*}[h!]
    \centering
    \renewcommand{\arraystretch}{1.1}
    \caption{Full results with the total numbers of trials and successful trials. The very right column (``avg.'') reports the average success rates over all tasks, which are already reported in the main paper. The other columns present a breakdown over different tasks. 
    }
    \resizebox{\textwidth}{!}{%
    \begin{tabular}{c@{\hspace{1ex}} l ccccc r}
        \toprule
        & & \multicolumn{5}{c}{Tasks} \\
        \cmidrule(rl){3-7}
        \textsc{\#} & Methods &
        \def\arraystretch{0.95}\begin{tabular}{@{}c@{}}boil water in\\the microwave\end{tabular} &
        \def\arraystretch{0.95}\begin{tabular}{@{}c@{}}bring in\\empty bottle\end{tabular} &
        \def\arraystretch{0.95}\begin{tabular}{@{}c@{}}cook a\\frozen pie\end{tabular} &
        \def\arraystretch{0.95}\begin{tabular}{@{}c@{}}halve\\an egg\end{tabular} &
        \def\arraystretch{0.95}\begin{tabular}{@{}c@{}}store\\firewood\end{tabular} & avg. (\%)\\
        \midrule
        \multicolumn{1}{l}{{\textbf{Ours}}} & \\
        \textsc{1} & \tpvqa{} (w/ GPT-4) & 12/18 & 18/20 & 14/23 & 15/20 & 8/20 & \bf 66.5\\

        \midrule
        \multicolumn{1}{l}{{\textbf{Baselines in Literature}}} & \\
        \textsc{2} & \llm{} & 1/20 & 2/34 & 0/20 & 0/19 & 0/22 & 2.2\\        
        \textsc{3} & \openloop{} & 5/35 & 3/11 & 4/30 & 8/30 & 1/25 & 17.1\\
        \textsc{4} & \affordance{} & 4/28 & 11/20 & 7/20 & 10/20 & 5/20 & 35.9\\
        \textsc{5} & \success{} & 7/18 & 18/20 & 10/20 & 12/20 & 4/20 & 51.8\\
        \textsc{6} & \successaffordance{} & 8/20 & 12/16 & 11/20 & 13/20 & 7/20 & 54.0\\
        
        \midrule
        \multicolumn{1}{l}{\textbf{{Ablations}}} & \\
        \textsc{7} & \effect{} & 15/30 & 15/16 & 4/15 & 10/15 & 7/25 & 53.0\\
        \textsc{8} & \precondition{} & 3/17 & 15/20 & 7/20 & 11/20 & 5/20 & 41.5\\

        \midrule
        \multicolumn{1}{l}{{\textbf{Other VLMs}}} & \\
        \textsc{9} & \tpvqa{} (w/ Gemini-1.5) & 7/20 & 14/20 & 6/20 & 9/20 & 8/20 & 44.0\\
        \textsc{10} & \tpvqa{} (w/ Claude-3) & 5/20 & 12/28 & 4/14 & 10/20 & 3/13 & 33.9\\

        \bottomrule
    \end{tabular}%
    }
\label{tab:bytask}
\end{table*}

\pagebreak

Next, we show Table~\ref{tab:params} that presents the parameters for specifying the \textbf{openness} of the simulation environments used for experiments. 
This table overlaps with Table~\ref{tab:situations} in the main paper on  the list of actions and situations. 
Beyond that, we present the probability of each individual situation taking place in the execution of the corresponding action. 
One can realize testing domains with different levels of openness by adjusting those probabilities. 
There are situations whose probabilities are out of our control, which are labeled ``N/A'' in the table. 
For instance, the occurrence of ``object is not inview'' depends on the robot's motion planner used for navigation. 
As a result, we are sure that there exist such situations in the experiments but the chance cannot be specified.

\begin{table}[h]
\centering
\caption{Actions and their situation parameters.}
\scriptsize
\renewcommand{\arraystretch}{0.9}
\begin{tabular}{@{}llr@{}}
\toprule
	  & \multicolumn{2}{c}{Uncertaint outcomes} \\
        \cmidrule(rl){2-3}
        Actions & 
        \def\arraystretch{0.95}\begin{tabular}{@{}c@{}}Situations\end{tabular} &
        \def\arraystretch{0.95}\begin{tabular}{@{}c@{}}Probabilities\end{tabular}\\ \midrule
	find & \begin{tabular}{@{}l@{}}(1) The robot succeeds in navigation but the object is not inview. \\(2) There is no free space near the object so navigation fails. \\(3) The object that the robot is holding drops during navigation. \end{tabular} & \begin{tabular}{@{}r@{}}N/A\\N/A\\0.1\end{tabular}\\ \midrule
	  grasp & \begin{tabular}{@{}l@{}}(1) The robot fails to grasp, and the object position remains unchanged.\\(2) The robot fails to grasp, and the object drops nearby. \end{tabular} & \begin{tabular}{@{}r@{}}0.25\\0.25\end{tabular}\\ \midrule
	  placein & \begin{tabular}{@{}l@{}}(1) The robot fails to place, and the object remains in the robot's hand.\\(2) The robot fails to place, and the object drops nearby. \end{tabular} & \begin{tabular}{@{}r@{}}0.1\\0.1\end{tabular}\\ \midrule
	  placeon & \begin{tabular}{@{}l@{}}(1) The robot fails to place, and the object remains in the robot's hand.\\(2) The robot fails to place, and the object drops nearby.\end{tabular} & \begin{tabular}{@{}r@{}}0.1\\0.1\end{tabular}\\ \midrule
	  fillsink & \begin{tabular}{@{}l@{}}(1) The robot fails to open the faucet. \end{tabular} &  \begin{tabular}{@{}r@{}}0.1\end{tabular}\\\midrule
        fill & \begin{tabular}{@{}l@{}}(1) The container is not fully filled.\\(2) The container drops nearby. \end{tabular} & \begin{tabular}{@{}r@{}}0.05\\0.05\end{tabular}\\ \midrule
        open & \begin{tabular}{@{}l@{}}(1) The robot fails to open and the object remains closed. \end{tabular} & \begin{tabular}{@{}r@{}}0.1\end{tabular}\\ \midrule
        close & \begin{tabular}{@{}l@{}}(1) The robot fails to close and the object remains open.\end{tabular} & \begin{tabular}{@{}r@{}}0.1\end{tabular}\\ \midrule
        turnon & \begin{tabular}{@{}l@{}}(1) The robot fails to turn on the switch and the object remains off.\end{tabular} & \begin{tabular}{@{}r@{}}0.1\end{tabular}\\ \midrule
        cut & \begin{tabular}{@{}l@{}}(1) The object is not cut into half, and the knife is still in the robot's hand.\\(2) The object is not cut into half, and the knife drops nearby.\end{tabular} & \begin{tabular}{@{}r@{}}0.25\\0.25\end{tabular}\\
\bottomrule
\label{tab:params}
\end{tabular}
\end{table}

Lastly, we present the checkpoints of the VLMs used in the experiments of this research. 
We tried our best to use the state-of-the-art VLMs when the experiments were conducted. 

\begin{table}[h]
\centering

\caption{Model checkpoints we used for off-the-shelf VLMs.}
\begin{tabular}{l|c}
\toprule
\textbf{VLM} & \textbf{Model} \\
\midrule
GPT-4 & gpt-4-turbo \\
\midrule
Gemini & gemini-1.5-pro \\
\midrule
Claude & claude-3-opus-20240229 \\
\bottomrule
\end{tabular}
\end{table}

\end{document}